\documentclass[letterpaper, 10 pt, conference]{ieeeconf}  

\IEEEoverridecommandlockouts                              

\usepackage{graphicx} 
\usepackage{subfigure}
\usepackage{algorithm} 

\usepackage{algpseudocode}  
\usepackage{amsmath} 
\usepackage{threeparttable}
\usepackage{CJK}
\usepackage{indentfirst}
\usepackage{amsmath}
\usepackage{cases}

\usepackage{amssymb}
\usepackage{bbding}
\usepackage{utfsym}
\usepackage{amsmath,amssymb}
\usepackage{mathrsfs}
\usepackage{multirow}
\usepackage{soul, color, xcolor}
\soulregister{\cite}7 
\soulregister{\citep}7 
\soulregister{\citet}7 
\soulregister{\ref}7 
\soulregister{\pageref}7 

\title{\LARGE \bf
Exploring Pose-Guided Imitation Learning for Robotic Precise Insertion
}

\author{Han Sun, Sheng Liu, Yizhao Wang, Zhenning Zhou, Shuai Wang, Haibo Yang, \\
Jingyuan Sun$^{\dagger}$, Qixin Cao   
	\thanks{Han Sun, Yizhao Wang, Zhenning Zhou and Qixin Cao  are with the School of Mechanical Engineering, Shanghai Jiao Tong University, Shanghai 200240 China. 
	}  
    \thanks{Sheng Liu is with Karlsruhe Institute of Technology, Karlsruhe, Germany.
	}  
	\thanks{Jingyuan Sun is with the Shanghai Huawei Technologies Co., Ltd., Shanghai 201799 China (corresponding author to provide phone: 18603680666; fax: none; e-mail:sunj549@gmail.com). } 
   }
\begin{document}

\maketitle
\thispagestyle{empty}
\pagestyle{empty}

\begin{abstract}
Imitation learning is promising for robotic manipulation, but \emph{precise insertion} in the real world remains difficult due to contact-rich dynamics, tight clearances, and limited demonstrations. Many existing visuomotor policies depend on high-dimensional RGB/point-cloud observations, which can be data-inefficient and generalize poorly under pose variations.
In this paper, we study pose-guided imitation learning by using object poses in $\mathrm{SE}(3)$ as compact, object-centric observations for precise insertion tasks.
First, we propose a diffusion policy for precise insertion that observes the \emph{relative} $\mathrm{SE}(3)$ pose of the source object with respect to the target object and predicts a future relative pose trajectory as its action.
Second, to improve robustness to pose estimation noise, we augment the pose-guided policy with RGBD cues. Specifically, we introduce a goal-conditioned RGBD encoder to capture the discrepancy between current and goal observations. We further propose a pose-guided residual gated fusion module, where pose features provide the primary control signal and RGBD features adaptively compensate when pose estimates are unreliable.
We evaluate our methods on six real-robot precise insertion tasks and achieve high performance with only $7$--$10$ demonstrations per task. In our setup, the proposed policies succeed on tasks with clearances down to $0.01$~mm and demonstrate improved data efficiency and generalization over existing baselines.
Code will be available at https://github.com/sunhan1997/PoseInsert.

\end{abstract}

\section{INTRODUCTION}

Robotic precise insertion in unstructured environments is a challenging manipulation problem that requires accurate perception and reliable closed-loop control under contact-rich dynamics and tight clearances \cite{10802423}.

\begin{figure}
	\centering
	\includegraphics[scale=0.57]{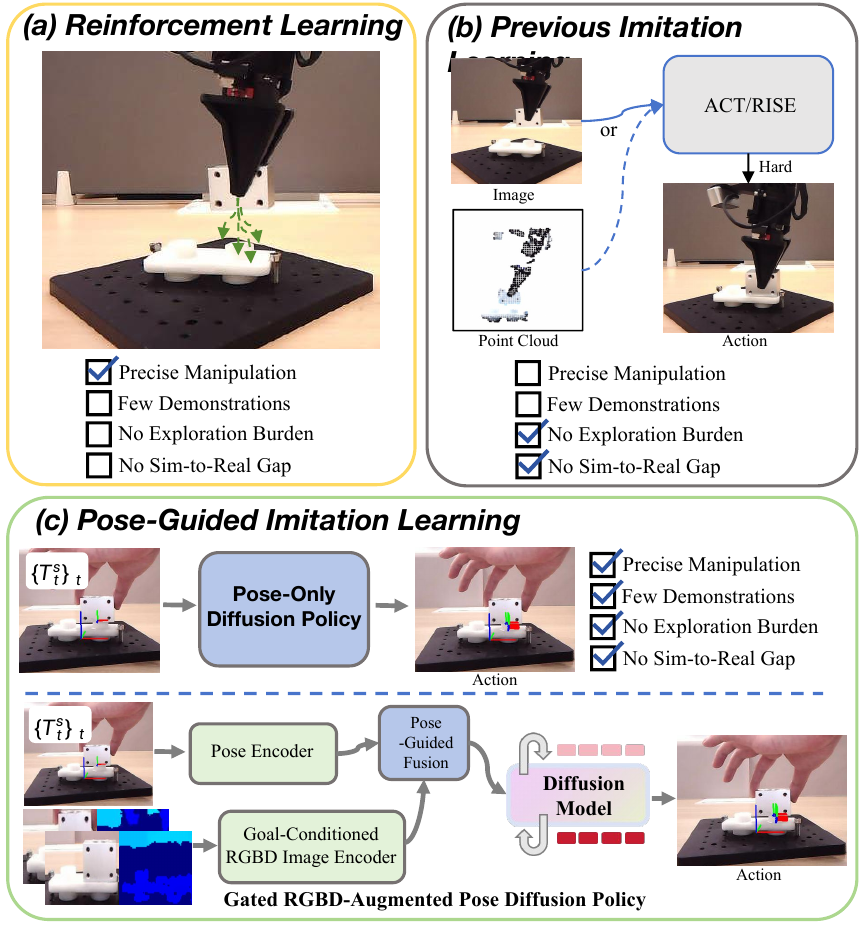}
	\caption{As depicted in (a), RL methods face challenges of inefficient exploration and a sim-to-real gap. Recent studies in (b) utilize the image/point cloud as input to learn action, which does not enable precise manipulation with few demonstrations. In contrast, our framework in (c) achieves precise manipulation with few demonstrations.
	}
	\vspace{-1.5em}
	\label{first}
\end{figure}

To acquire closed-loop insertion behaviors, prior works \cite{dong2021tactile, schoettlerdeep, beltran2020learning, ma2024automated,schoettler2020meta} combine reinforcement learning (RL) with force or tactile feedback. While effective in some settings, these methods often require substantial exploration, suffer from sparse rewards and unstable training \cite{mnih2015human}, and face non-trivial sim-to-real transfer (see Fig.~\ref{first}). Generalizing RL policies across varying object poses and spatial configurations also remains difficult.

In contrast, imitation learning is often more practical for real-world robotics: it leverages supervised learning from demonstrations and reduces reward design and exploration burdens. Recent studies \cite{zhao2023learning, chi2023diffusion, 10801678, goyal2023rvt} train end-to-end visuomotor policies from RGB images or point clouds (Fig.~\ref{first}). However, these high-dimensional observations typically require dozens to hundreds of demonstrations \cite{zhao2023learning,mandlekar2023mimicgen, kim2024openvla} to cover viewpoint changes and out-of-distribution pose configurations, which is especially costly for \emph{precise} manipulation.

To improve data efficiency, object-centric representations have been explored, including object detection \cite{zhu2023viola}, 6D object poses \cite{wen2022you, hsu2024spot}, and 2D/3D flow \cite{xu2024flow, yuan2024general, zhu2024vision}. These structured signals provide compact, task-relevant inputs for downstream policies. Nevertheless, many methods still degrade under significant pose disturbances, and pose-guided policies can be brittle when online pose estimation is noisy---a common failure mode for small connectors and tight clearances.

Existing learning-based approaches therefore still struggle to achieve \emph{precise} insertion that is both data-efficient and robust to pose variations. For example, \cite{wen2022you} studies category-level behavior cloning for insertion but relies on key-pose triggering and specialized sensing, while \cite{papagiannis2024miles} learns contact-rich manipulation from a single demonstration but generalizes poorly to unseen pose configurations. Therefore, developing an imitation learning method that can (i) learn precise insertion from few demonstrations and (ii) generalize under pose disturbances remains a significant challenge.

In this work, we explore pose-guided imitation learning for precise insertion by using object poses in $\mathrm{SE}(3)$ as compact, object-centric observations instead of raw RGB or point clouds. We study two complementary designs:
\textbf{(1)} a pose-only policy that takes the relative $\mathrm{SE}(3)$ pose of the source object with respect to the target object as observation and predicts a future relative pose trajectory as action; and
\textbf{(2)} an RGBD-augmented policy that uses goal-conditioned visual cues to compensate for pose estimation noise via an adaptive gated fusion mechanism.
Our approach emphasizes perception-guided trajectory generation and does \emph{not} require force/torque sensing; instead, we execute policies on an ALOHA-like robot that provides passive compliance during contact. Force/tactile feedback is complementary and will be discussed as a promising direction for more challenging contact regimes.

In summary, our contributions are threefold:
\begin{itemize}
    \item We propose a pose-guided imitation learning framework for robotic precise insertion, featuring a diffusion policy that predicts future relative $\mathrm{SE}(3)$ pose trajectories. We introduce a disentangled pose encoder to improve representation learning for complex trajectory prediction.
    \item To mitigate pose estimation noise, we augment the pose-guided diffusion policy with RGBD observations via a goal-conditioned RGBD encoder and a pose-guided residual gated fusion module, where pose features form the backbone and RGBD cues provide adaptive residual corrections.
    \item We demonstrate strong generalization and robustness on six real-robot precise insertion tasks with only $7$--$10$ demonstrations per task, including tight-clearance insertion in our setup.
\end{itemize}

\section{RELATED WORK}

\subsection{Robotic Precise Insertion}
Robotic precise insertion is a long-standing challenge due to tight clearances, contact-rich dynamics, and the need for reliable closed-loop control \cite{10802423, mason2018toward, thomas2018learning}. 
Many learning-based approaches incorporate additional contact sensing to stabilize insertion. For instance, \cite{dong2021tactile} studies tactile-based RL for part insertion, while \cite{schoettlerdeep, beltran2020learning, ma2024automated} combine RL with force/torque feedback for contact-rich assembly.
Despite their effectiveness in some settings, these methods often require substantial exploration, can suffer from sparse rewards and unstable training, and face sim-to-real challenges when trained in simulation. Generalization across varying object poses and unseen spatial configurations also remains difficult, motivating data-efficient methods that are robust to pose disturbances in real-world insertion.

\subsection{End-to-end Imitation Learning}
Imitation learning has achieved strong performance on complex manipulation by mapping raw sensory observations to robot actions \cite{chi2023diffusion, 10801678, zhao2023learning, shridhar2023perceiver}. 
ACT \cite{zhao2023learning} uses transformer-based policies with image encoders, while Diffusion Policy \cite{chi2023diffusion} models multi-modal action distributions via a diffusion process.
Several works incorporate 3D observations to improve geometric reasoning \cite{10801678, shridhar2023perceiver, ze20243d}; for example, DP3 \cite{ze20243d} leverages 3D perception in manipulation policies, and RISE \cite{10801678} predicts actions from point clouds using a sparse 3D encoder.
However, these end-to-end approaches typically rely on high-dimensional observations (RGB or point clouds) and thus require large-scale demonstrations to cover viewpoint and pose variations, making them particularly data-inefficient for precise insertion. This motivates compact, geometry-aware representations that better capture contact-rich alignment.

\subsection{Object-Centric Imitation Learning}
To improve efficiency and generalization, object-centric methods extract structured information from visual observations---such as object detection \cite{zhu2023viola}, point tracking \cite{huang2024rekep}, 6D object poses \cite{wen2022you, hsu2024spot}, and 2D/3D flow \cite{xu2024flow, yuan2024general, zhu2024vision}---and use them as inputs for downstream policies.
Among these representations, 6D object pose provides a compact geometric description that can reduce policy learning complexity and facilitate generalization across spatial configurations.
Pose features are also less sensitive to appearance changes (e.g., background clutter and illumination) \emph{when pose estimation is reliable}; however, robustness still depends on online pose estimation quality, and failures under occlusion or for small objects remain practical concerns.

In this paper, we build on the object-centric perspective and study pose-guided imitation learning for robotic precise insertion.
Different from end-to-end visuomotor policies, we use the \emph{relative} $\mathrm{SE}(3)$ pose between the source and target objects as a compact observation--action interface, and further introduce goal-conditioned RGBD cues to compensate for pose estimation noise during execution.

\section{PROPOSED APPROACH}
In this section, we present a detailed description of the proposed pose-only diffusion policy and gated RGBD-augmented pose diffusion policy. Additionally, we introduce the corresponding human demonstration data collection methods tailored for each approach.

\textbf{Notation.} We denote a rigid transform from frame $a$ to frame $b$ as $T_a^b \in SE(3)$, consisting of rotation $R_a^b \in SO(3)$ and translation $t_a^b \in \mathbb{R}^3$. 
We use $c$ for the camera frame, $b$ for the robot base, and $e$ for the end-effector. 
$T_c^s$ and $T_c^t$ denote the estimated poses of the source/target objects in the camera frame.
The relative pose used in this work is $T_t^s = (T_c^t)^{-1}T_c^s$.

\begin{figure}[t]
	\centering
	\hspace{-1.2em}
	\subfigure[ ]
	{
		\begin{minipage}[b]{.97\linewidth}
			\centering
			\includegraphics[scale=0.38]{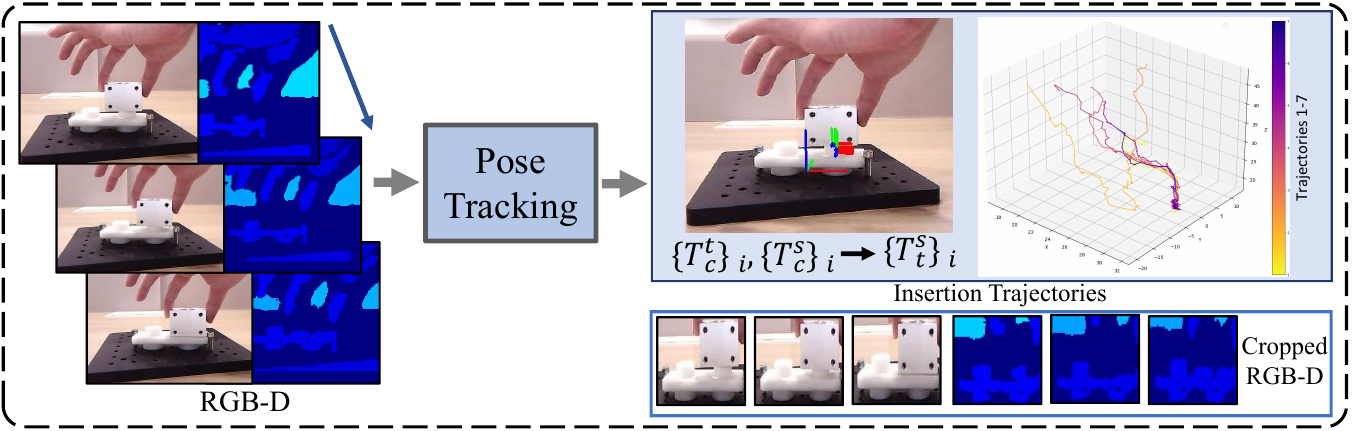}
		\end{minipage}
	}
	
	\hspace{-1.2em}
	\subfigure[]
	{
		\begin{minipage}[b]{.97\linewidth}
			\centering
			\includegraphics[scale=0.38]{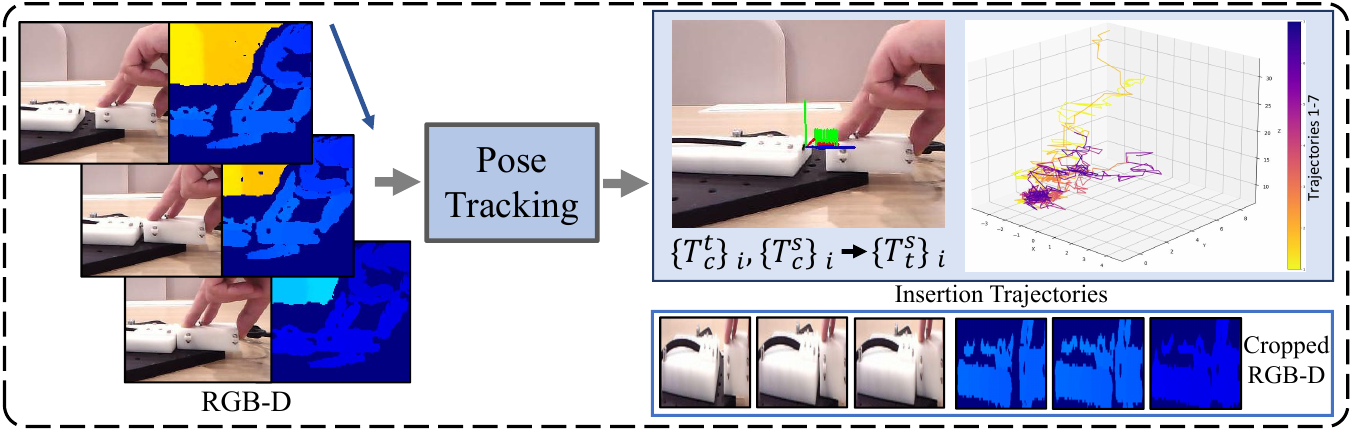}
		\end{minipage}
	}
	\vspace{-0.8em}
	\caption{ Human demonstration data collection. (a) Data collection and pose trajectory visualization for metal part insertion task. (b) Data collection and pose trajectory visualization for USB (Type-C) insertion task. (Trajectory Unit: Millimeter)  }
	\label{data}
\end{figure}

\subsection{Pose-Only Diffusion Policy}
\label{Pure}
In this paper, we opt for 6D object pose as the object-centric representation, which depicts the full 6D state information of rigid objects throughout the episode. The insertion task involves a set of objects $v = \left \{ v^{s}, v^{t} \right \} $, consisting of a graspable source object $v^{s}$ and a target object $v^{t}$. In practice, we apply FoundationPose \cite{wen2024foundationpose}, a zero-shot 6D object pose estimation and tracking method, to extract the poses for both the source and target objects.

\textbf{Human Demonstration Data Collection.}
The source object pose in the camera frame is defined as  $T_{c}^{s}$, the target object pose in the camera frame is defined as  $T_{c}^{t}$. As shown in Fig.~\ref{data}, given the human demonstration video, the goal is to extract pose trajectories $\left \{T_{c}^{s}\right \}_{i}$ and $\left \{T_{c}^{t}\right \}_{i}$, corresponding to object $v^{s}$ and $v^{t}$, respectively. 
We then convert all source object poses into the target object’s coordinate frame and obtain the relative pose trajectory $ \hat{\tau} =  \left \{T_{t}^{s} \right \}_{i} $. This transforms multiple demonstration trajectories into a canonicalized space, which ignores the absolute configurations and enhances the generalization ability of the model. In addition, these trajectories describe the object-centric motions in 3D and thus implicitly encode the search and contact constraints throughout the insertion task.  
Note that the pose trajectories are obtained from the estimator and thus serve as pseudo ground-truth supervision; pose noise may affect learning, motivating the RGBD augmentation in Section~\ref{RGBD}.

\begin{figure*}
	\centering
	\includegraphics[width=1.0\linewidth]{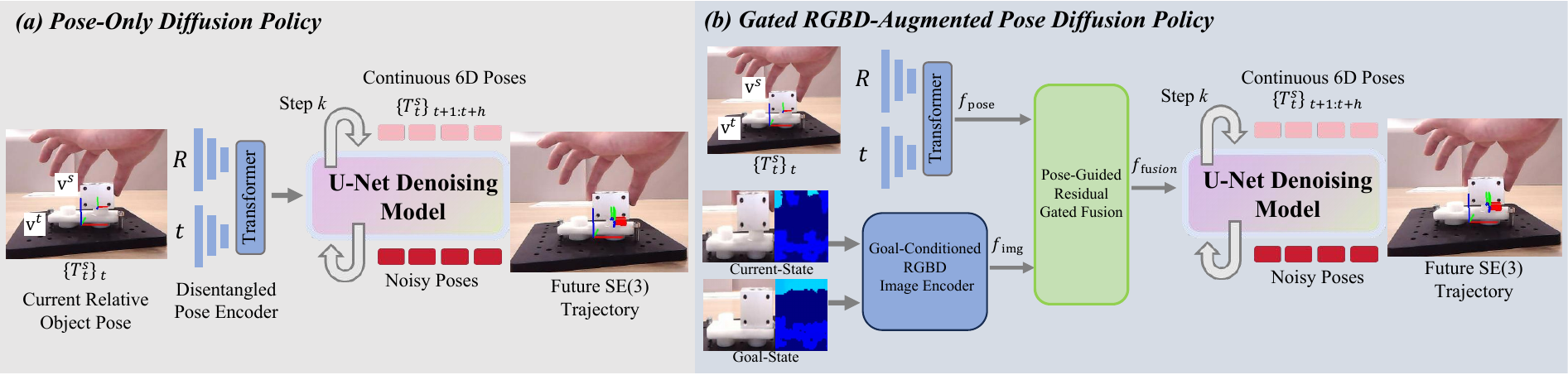}
	\caption{\textbf{Pipeline of the pose-guided imitation learning methods.} \textbf{(a)} For the \textbf{pose-only diffusion policy}, current source object pose relative to the target object $\left \{T_{t}^{s} \right \}_{t}$ is considered as observation. The disentangled pose encoder is applied to extract the pose features. Then, diffusion policy predicts the future relative $SE(3)$ trajectory $\left \{T_{t}^{s} \right \}_{t+1:t+h} $. \textbf{(b)} For the \textbf{gated RGBD-augmented pose diffusion policy}, current relative object pose $\left \{T_{t}^{s} \right \}_{t}$ is fed to the disentangled pose encoder to get the $f_{pose}$, the current RGBD image patch is sent to goal-conditioned RGBD image encoder to obtain the $f_{img}$. And the residual gated fusion module uses image features to assist pose features. The enhanced fusion feature $f_{fusion}$ is fed to diffusion policy to predict the future relative $SE(3)$ trajectory $\left \{T_{t}^{s} \right \}_{t+1:t+h} $. }
	\label{pipeline}
\end{figure*}

\begin{figure}[t]
	\centering
	\subfigure[ ]
	{
		\begin{minipage}[b]{.6\linewidth}
			\centering
			\includegraphics[scale=0.38]{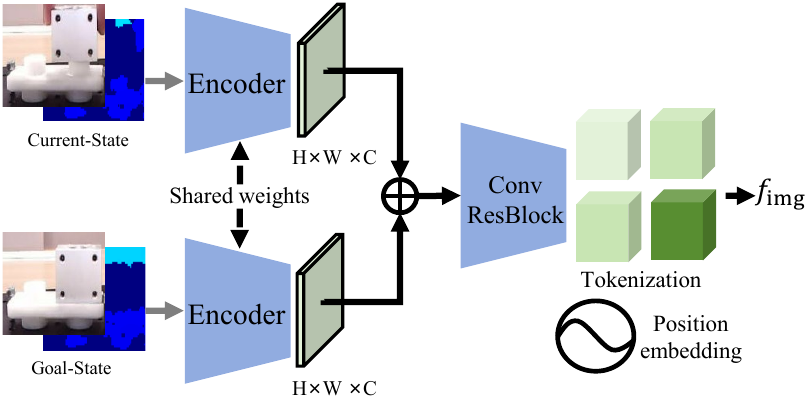}
		\end{minipage}
	}
    \hspace{-1em}
	\subfigure[]
	{
		\begin{minipage}[b]{.35\linewidth}
			\centering
			\includegraphics[scale=0.43]{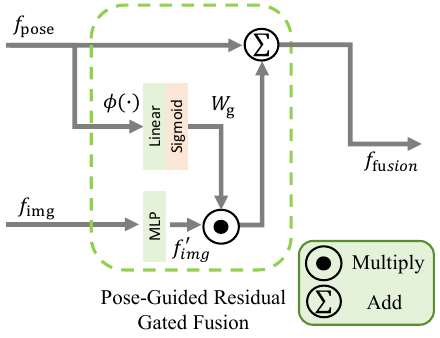}
		\end{minipage}
	}
	\caption{ (a) Goal-conditioned RGBD image encoder. We adopt a RefineNet from \cite{wen2024foundationpose} to extract the RGBD features.  The two branch inputs of the refinenet are changed to the current RGBD image patch and the goal RGBD image patch, respectively.   (b) Pose-guided residual gated fusion module. The MLP layer consists of a linear transformation, LayerNorm, and ReLU activation. }
	\label{image_encoder1}
\end{figure}

\textbf{Disentangled Pose Encoder.}
As shown in Fig.~\ref{pipeline}, unlike \cite{hsu2024spot}, a disentangled pose encoder is designed to extract the SE(3) pose features. Inspired by \cite{li2019cdpn, wang2021gdr}, we design a translation network and rotation network to extract the features of translation $ t \in \mathbb{R}^{3} $ and rotation $ R \in \mathbb{R}^{3 \times 3} $ respectively. The translation network employs a three-layer MLP, LayerNorm layers, and a GELU activation function, which encodes $t$ into a 64-dimensional vector $F_{t}$. Similarly, the rotation network encodes $R$ into a 64-dimensional vector $F_{r}$. Then, $F_{t}$ and  $F_{r}$ are concatenated to form a 128-dimensional vector, which is then fed to a Transformer layer to capture the implicit correlation between pose components through the self-attention mechanism and finally outputs the 128-dimensional pose features $F_{pose}$ by linear projection.

\textbf{Diffusion Policy.}
 In general, imitation learning methods \cite{10801678,ze20243d} model $p(\mathbf{A}_{t}|\mathbf{O}_{t})$  to infer a sequence of future actions $\mathbf{A}_{t}$ from the current observations $\mathbf{O}_{t}$. To adapt this method to our task, we define the observation input $\mathbf{O}_{t}$ as the current relative pose  $\left \{T_{t}^{s} \right \}_{t} \in SE(3)$. The action output $\mathbf{A}_{t}$ is defined as the future relative poses over a prediction horizon of $h$ time steps:  $\left \{T_{t}^{s} \right \}_{t+1:t+h} \in SE(3)^{h}$, where $h$ represents the number of future time steps to be predicted.


\begin{algorithm}[] 
	\renewcommand{\algorithmicrequire}{\textbf{Input:}}
	\renewcommand{\algorithmicensure}{\textbf{Output:}}
	\caption{Pose Diffusion Policy Inference}  
	\label{alg:pose_diffusion_policy}  
	\begin{algorithmic}[1]  
		\Require  
		
		{End-effector pose in base frame: $T_{b}^{e} $; Camera pose in base frame: $T_{b}^{c}$; $T_{c}^{s}$;  $T_{c}^{t}$;}
		\Ensure  
		End-effector pose Trajectory: $\{T_{b}^{e}\}_{t=0}^{N}$ in base frame
		\For{$t \gets 0$ to $N$}  
		\State Relative object pose: 
		$T_{t}^{s} \gets (T_{c}^{t})^{-1} \cdot T_{c}^{s}$  
		
		\State Source object to base frame: 
		$T_{b}^{s} \gets T_{b}^{c} \cdot T_{c}^{s}$  
		
		\State End-effector in source object: 
		$T_{s}^{e} \gets  (T_{b}^{s})^{-1} \cdot T_{b}^{e} $
		
		\State Predict relative pose trajectories:
		$\{T_{t}^{s}\}_{h} \gets \text{DP}(T_{t}^{s})$  
		
		\Comment{DP=Diffusion Policy}
		
		\State Transform trajectory to camera frame: 
		
		$\{T_{c}^{s}\}_{h} \gets T_{c}^{t} \cdot \{T_{t}^{s}\}_{h}  $
		
		\State Transform trajectory to base frame: 
		
		$\{T_{b}^{s}\}_{h} \gets T_{b}^{c} \cdot \{T_{c}^{s}\}_{h}$
		
		\State Compute final end-effector trajectory: 
		
		$\{T_{b}^{e}\}_{h} \gets  \{T_{b}^{s}\}_{h}  \cdot T_{s}^{e}$
		
		\State Execute the end-effector action: $\{T_{b}^{e}\}_{n}$
		\State Add $\{T_{b}^{e}\}_{n}$ to $\{T_{b}^{e}\}_{t=0}^{N} $
		
		\EndFor  
	\end{algorithmic}  
				\vspace{-0.2em}
\end{algorithm}

Our method is built upon a diffusion policy, which leverages a conditional denoising diffusion model conditioned on the observation pose $\mathbf{O}_{t}$, which denoises Gaussian noise noise into actions $\mathbf{A}_{t}$. Specifically, starting from a Gaussian noise $\mathbf{A}_{t}^{k}$, the denoising network $\varepsilon_{\theta }$ performs $K$ iterations to gradually denoise a random noise $A_{t}^{k}$ into the noise-free action $\mathbf{A}_{t}^{0}$,
\begin{equation}
	A_{t}^{k-1} = \alpha_{k}  (A_{t}^{k} -   \gamma_{k} \varepsilon_{\theta }(O_{t},
	A_{t}^{k} ,k) +   \sigma_{k}  \mathcal{N} (0, I )  )
\end{equation}
where $\alpha_{k}$, $\gamma_{k}$ and $\sigma_{k}$ are hyperparameters associated with step k in the noise schedule. This process is also called the reverse process \cite{ho2020denoising}. Mean squared error (MSE) loss is employed as the objective function to supervise the prediction of object pose:
\begin{equation}
\mathcal{ L} = MSE(\epsilon^{k}, \varepsilon _{\theta }  (O_{t}, \hat{\alpha}_{k}A_{t}^{0}+ \hat{\beta}_{k}\epsilon^{k},k) )
\end{equation}
where $\epsilon^{k}$ is the noise at iteration $k$, $\hat{\alpha}_{k}$ and  $\hat{\beta}_{k}$ are noise-schedule coefficients used for the one-step forward noising process \cite{ho2020denoising}. The rotation is represented as R6D \cite{wang2021gdr}, a continuous 6D representation for $SO(3)$ that avoids discontinuities (e.g., quaternion sign ambiguity) and is empirically stable for regression and generative modeling.
We use a diffusion policy with a convolutional network (U-Net). DDIM \cite{song2020denoising} is utilized as the noise scheduler, and we use sample prediction instead of epsilon prediction.   

\textbf{End-effector Action Generation.}
At each timestep, the policy predicts a desired relative pose $T_{t}^{s,\mathrm{des}}$ in the target frame. We convert it to a desired source pose in the camera frame via
$T_{c}^{s,\mathrm{des}} = T_c^t\, T_{t}^{s,\mathrm{des}}$,
and then into the robot base frame using the calibrated $T_b^c$.
The end-effector command is computed using the (estimated) grasp transform between the gripper and the source object.
This receding-horizon closed-loop execution uses the \emph{relative} object geometry as feedback, which reduces sensitivity to global placement errors. Full transformation details are provided in Alg.~\ref{alg:pose_diffusion_policy}.

\subsection{Gated RGBD-Augmented Pose Diffusion Policy}
\label{RGBD}
In Section \ref{Pure}, the proposed pose-only diffusion policy learns the SE(3) pose trajectory of the source object relative to the target object, effectively modeling complex insertion trajectories. This method relies on the accuracy of 6D pose estimation. As shown in Fig.~\ref{data} (a), when the pose estimation algorithm provides precise 6D poses for both the source and target objects, the method achieves strong performance and demonstrates impressive generalization across diverse scenarios, which is shown in Section \ref{Pure_exp}. However,  as shown in Fig.~\ref{data} (b), under noisy pose conditions, the model may struggle to learn meaningful trajectories from demonstrations. To address this limitation, we propose augmenting the framework with RGBD information to dynamically correct pose errors and enhance trajectory robustness.

\textbf{Human Demonstration Data Collection.}  
Similar to Section \ref{Pure}, in addition to the relative pose trajectory $\hat{\tau} = \{T_{t}^{s}\}_i$, we capture RGBD images during insertion. The 6D poses of source/target objects are estimated, along with their axis-aligned bounding boxes:
\begin{equation}
B_s=((x_{s}^{1}, y_{s}^{1}), (x_{s}^{2}, y_{s}^{2})), 
B_t=((x_{t}^{1}, y_{t}^{1}), (x_{t}^{2}, y_{t}^{2}))
\end{equation}
The union bounding box is computed as:
\begin{equation}
B =  ((min(X), min(Y)),(max(X),  max(Y) ))
\end{equation}
where $X = \{x_{s}^{1}, x_{s}^{2}, x_{t}^{1}, x_{t}^{2}\}$ and $Y = \{y_{s}^{1}, y_{s}^{2}, y_{t}^{1}, y_{t}^{2}\}$. This ensures that the cropped RGBD patches $\{I\}_i$ encapsulate both objects, enhancing model generalization.

\textbf{Goal-Conditioned RGBD Image Encoder.}  
Inspired by \cite{wen2024foundationpose} and \cite{kim2024goal}, we propose a goal-conditioned RGBD image encoder to capture the discrepancy between current and goal states, providing visual error signals when pose estimates are noisy. Both the current $I_{c}$ and goal $I_{g}$
observations are cropped and resized to $320×320$ before being sent to the network. As shown in Fig.~\ref{image_encoder1} (a), the goal observation $I_{g}$ is explicitly fed into the network through channel-wise concatenation with the current observation $I_{c}$, enabling direct learning of the current-to-goal discrepancy. Following the RefineNet architecture in FoundationPose \cite{wen2024foundationpose}, we employ a shared CNN encoder to extract and align low-level features (e.g., edges, geometric structures) from both $I_{c}$ and $I_{g}$. The concatenated features are then processed by residual blocks \cite{he2016deep} and tokenized into patches with position embeddings \cite{dosovitskiy2020image}. This architecture facilitates the extraction of distinctive feature representations that capture the discrepancy between current and goal states, while simultaneously reducing computational overhead.

\textbf{Pose-Guided Residual Gated Fusion.}  
The introduction of RGBD observations aims to mitigate the influence of pose estimation noise, thereby refining the motion prediction of the source object. To achieve this, we propose a pose-guided residual gated fusion module.
In this module, the pose features serve as the primary features, while the gating weights for image features are dynamically computed based on the pose features. Through the gating mechanism, image features adaptively compensate for the limitations of pose features, ensuring a robust fusion of multimodal information. This design ensures that visual cues intervene dynamically based on pose features.

The pose features from  the disentangled pose encoder are defined as $f_{pose} \in \mathbb{R}^{B \times 128} $. As shown in Fig.~\ref{image_encoder1} (b), the image features from  the goal-conditioned RGBD image encoder are defined as $f_{img} \in \mathbb{R}^{B \times 1200} $.  The image features are sent to  a MLP layer,  which compresses $f_{img}$ to the same dimension  $f_{img}^{'} \in \mathbb{R}^{B \times 128} $ as the pose features. The gate layer $\phi(\cdot)$ consists of a linear transformation followed by a Sigmoid activation function. The pose features are processed through the gate layer to generate gating weights $W_{g}$,
\begin{equation}
    W_{g} = \phi(f_{pose})
\end{equation}
which enables the model to dynamically modulate visual feature contributions based on the current pose state through adaptive feature selection. The gating weights  $W_{g}$ are element-wise multiplied with the image features $f_{img}^{'} $, followed by residual connection fusion with the pose features  $f_{pose}$.
\begin{equation}
	f_{fusion} = f_{pose} +  W_{g}  \odot  f_{img}^{'} 
\end{equation}
where $\odot$  represents element-wise multiplication.  This architecture preserves the dominance of geometric priors while enabling conditional feature integration. The gating mechanism facilitates finer-grained cross-modal information flow control compared to conventional concatenation or additive operations. Crucially, using pose features as control signals aligns with the geometry-driven characteristics inherent in insertion tasks.


\textbf{Diffusion Policy.}
Similar to Section \ref{Pure}, we use the diffusion process to generate the SE(3) pose trajectory of the source object relative to the target object. The observation input $\mathbf{O}_{t}$ is defined as the current SE(3) relative pose  $\left \{T_{t}^{s} \right \}_{t} \in SE(3)$ and RGBD image patch  $\left \{I \right \}_{t} \in \mathbb{R}^{H \times W \times 4 }$.
The action output $\mathbf{A}_{t}$ is  the future relative pose over a prediction horizon of $h$ time steps:  $\left \{T_{t}^{s} \right \}_{t+1:t+h} \in SE(3)^{h}$. The loss function for diffusion policy and end-effector action generation follows the same formulation as in Section \ref{Pure}.

\section{EXPERIMENTS}
In this section,  we provide our implementation details and experimental setup, evaluate the proposed methods and present results compared with the state-of-the-art imitation learning methods in various real-world experiments.
Then, we conduct ablation studies to evaluate the proposed modules, including the Disentangled Pose Encoder (DPE), Goal-Conditioned RGBD Image Encoder  (GIE) and Pose-Guided Residual Gated Fusion  (PRGF). 
Furthermore, we provide further analysis.

\subsection{Experimental Setup}
\begin{figure}
	\centering
		\subfigure[]
	{
		\begin{minipage}[b]{.29\linewidth}
			\centering
			\includegraphics[scale=0.195]{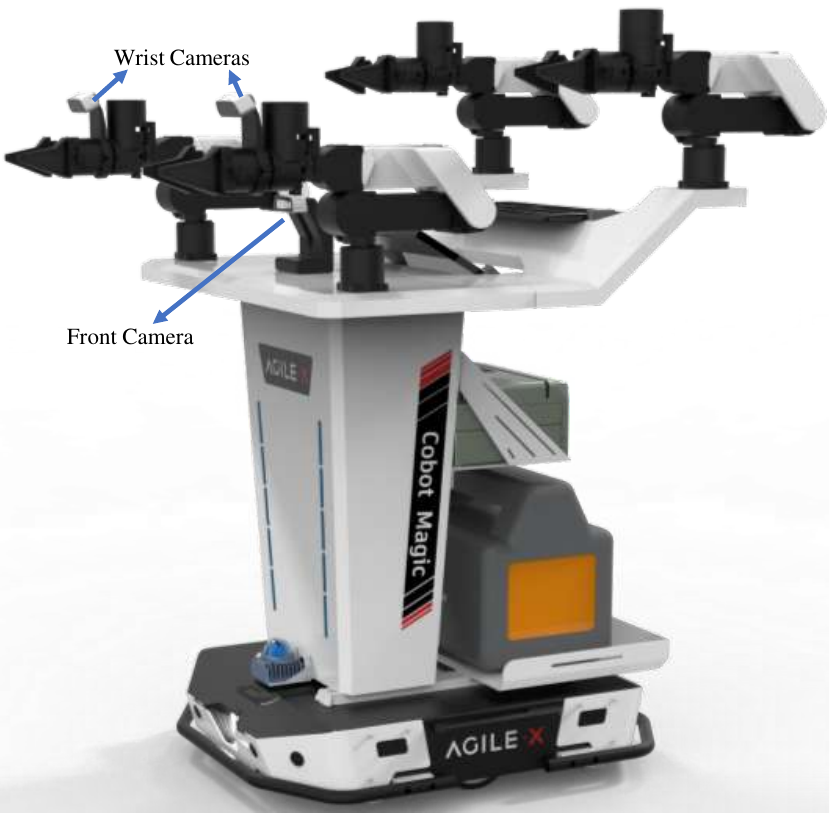}
		\end{minipage}
	}
	\hspace{-0.6em}
		\subfigure[ ]
	{
		\begin{minipage}[b]{.65\linewidth}
			\centering
			\includegraphics[scale=0.20]{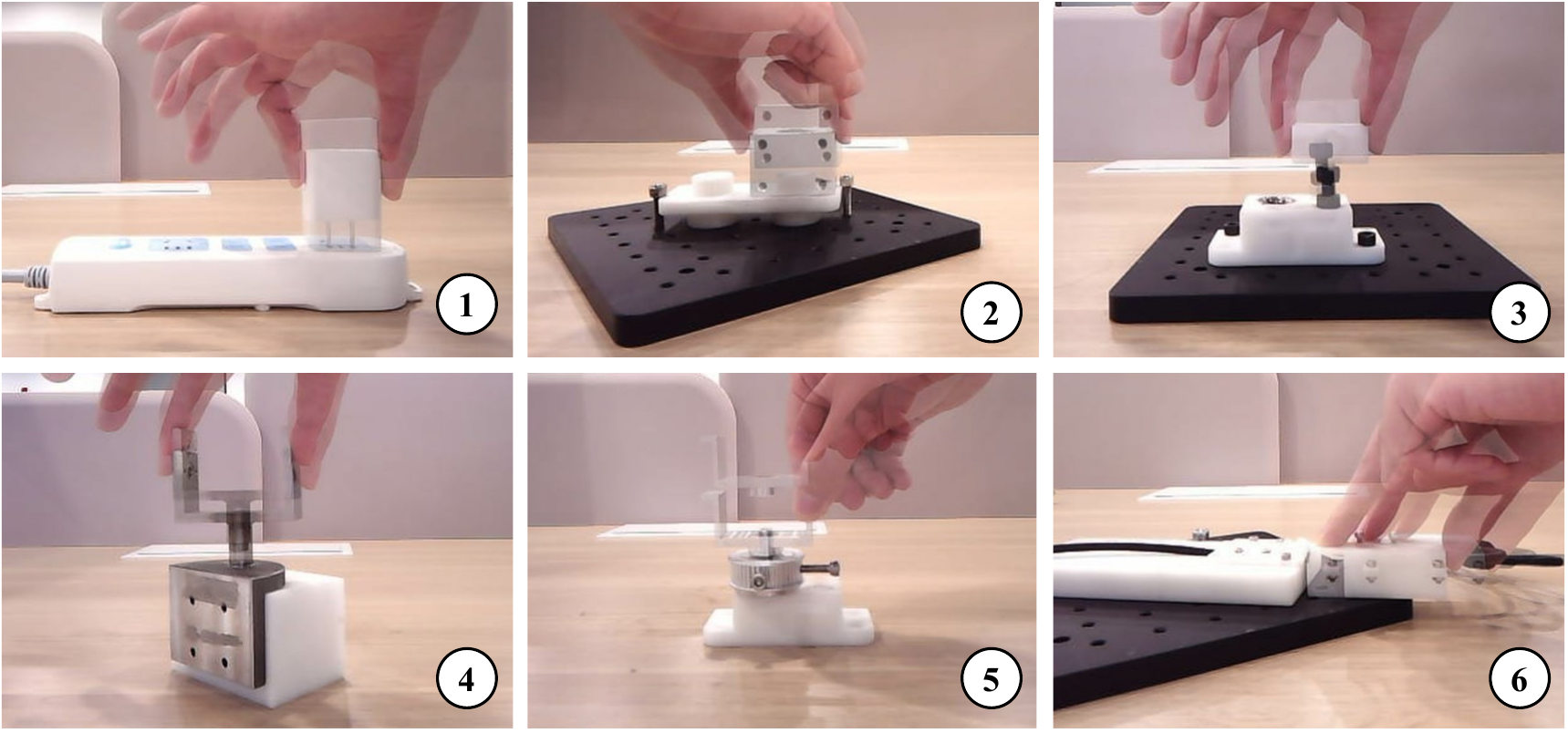}
		\end{minipage}
	}

	\caption{(a) Cobot Mobile ALOHA for experiments. (b) There are $6$ precise insertion tasks for the real-world experiments.
	}
	\label{exp_set}
\end{figure}

\textbf{Implementation Details.} All experiments are performed on the Cobot Mobile ALOHA, a robot using the Mobile ALOHA system design \cite{zhao2023learning} and manufactured by agilex.ai (Fig. \ref{exp_set}(a)). It is important to note that although this robot is not equipped with force sensors, it exhibits passive compliance during task execution, which means that it can perform contact-rich tasks safely. An Orbbec Dabai camera, the front camera in Fig.~\ref{exp_set}(a), is used as a vision sensor to observe the objects on the table.
We collect $7-10$ demonstrations for each task. During demonstration collection, we keep the target object placement fixed to reduce pose estimation failures and to obtain cleaner supervisory trajectories. During evaluation, we perturb the initial object placements to form in-distribution and out-of-distribution test conditions. All data are collected at $30$ Hz. All the experiments are conducted on a single NVIDIA RTX 4060 GPU. We train the networks using the AdamW optimizer with a batch size of $80$ and an initial learning rate of $3e-4$. We train for 2000 epochs.

\textbf{Task Description.}  As illustrated in Fig.~\ref{exp_set} (b), we carefully design $6$ precise insertion tasks, including (1) plug insertion, (2–5) metal-part insertion into a base, and (6) USB Type-C insertion. The measured clearances are $0.28$ mm for task 2, $0.08$ mm for task 3, $0.03$ mm for task 4, and $0.01$ mm for task 5. The difficulty of tasks $1$ to $6$ gradually increases.

\textbf{Evaluation Metrics.} We employ the success rate (SR) as the evaluation metric, which is calculated by dividing the number of successful trials by the total number of trials. Each policy is tested for $10$ consecutive trials to evaluate its performance.

\textbf{Baselines.}  We compare against the state-of-the-art imitation learning methods: ACT \cite{zhao2023learning}, RISE \cite{10801678} and SPOT \cite{hsu2024spot}. 
For ACT and RISE, we collect $50$ demonstrations per task following their typical data requirements for stable training on contact-rich behaviors. 
Since SPOT is not open-source, we re-implement it using an MLP encoder and a diffusion-policy backbone, and collect $10$ demonstrations per task for training to match the data regime used by our methods.

\begin{figure*}
	\centering
	
	\includegraphics[width=0.95\linewidth]{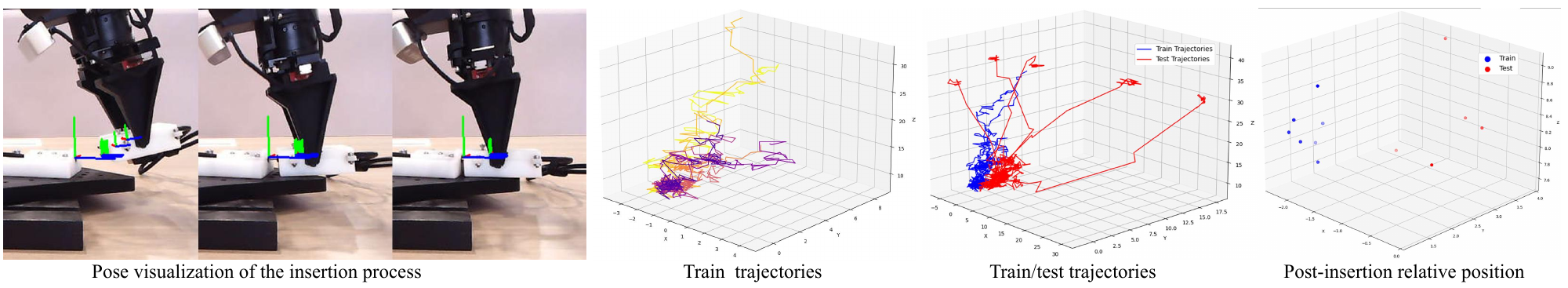}
	\caption{Visualization of train/test trajectories for task 6 (Unit: Millimeter).  }
	\vspace{-1.em}
	\label{vis}
\end{figure*}


\begin{table}[t]
	\caption{THE SUCCESS RATE (\%) OF OUR METHODS AND BASELINES. }
	\centering
			\vspace{-0.8em}
	\label{in_d}
		\setlength{\tabcolsep}{1.8mm}{
			\begin{tabular}{cccccccc}
				\hline
				\textbf{Methods}   & \textbf{1}     & \textbf{2}    & \textbf{3}     & \textbf{ 4}     & \textbf{5}     & \textbf{6}     & \textbf{Average}   \\ \hline
				ACT \cite{zhao2023learning}    &10    &0  &0  &0  &0  &0   &1.7 \\
				RISE \cite{10801678}       &20    &0  &0  &0  &0  &0    &3.3 \\
				SPOT \cite{hsu2024spot}        &\textbf{100}    &70  &\textbf{90}  &90  &\textbf{70}  &10   &71.7 \\ \hline
				PoseDP                         & \textbf{100}   &\textbf{100} &\textbf{90}  &\textbf{100} & \textbf{70}  & 30  &81.7  \\
				RPDP                 & \textbf{100}  &\textbf{100} &\textbf{90} &90  & \textbf{70}  & \textbf{100} &\textbf{91.7} \\ \hline
		\end{tabular} }
			\vspace{-1.2em}
	\end{table}

	\begin{table}[t]
		\caption{THE SUCCESS RATE (\%) OF OUR METHODS AND BASELINES. (OUT OF DISTRIBUTION)}
		\centering
				\vspace{-1.2em}
		\label{out_d}
			\setlength{\tabcolsep}{1.8mm}{
				\begin{tabular}{cccccccc}
					\hline
					\textbf{Methods}   & \textbf{1}     & \textbf{2}    & \textbf{3}     & \textbf{ 4}     & \textbf{5}     & \textbf{6}     & \textbf{Average}   \\ \hline
					ACT \cite{zhao2023learning}    &0    &0  &0  &0  &0  &0   &0 \\
					RISE \cite{10801678}       &0    &0  &0  &0  &0  &0   &0 \\
					SPOT \cite{hsu2024spot}        &90     &70  &80  &70  &50   &10  &61.7 \\ \hline
					PoseDP                         & \textbf{90}   &\textbf{100} &90  &\textbf{90} &60  & 30  &76.7 \\
					RPDP                 & 80  &90 &70  &\textbf{90}  &60  & \textbf{80} &\textbf{78.3} \\ \hline
			\end{tabular} }
				\vspace{-1.2em}
		\end{table}

\subsection{Pose-Only Diffusion Policy}
\label{Pure_exp}
Table \ref{in_d} presents the quantitative evaluation comparing the pose-only diffusion policy (PoseDP), Gated RGBD-Augmented Pose Diffusion Policy (RPDP)  with baselines when objects are positioned within the train distribution range, while Table \ref{out_d} demonstrates the comparative results for out-of-distribution conditions. 

As shown in Table \ref{in_d}, the ACT and RISE failed in all $10$ trials across the tasks $2$ to $6$. Due to the larger insertion clearance in task $1$, ACT achieved one successful insertion, while RISE succeeded twice.  Table \ref{out_d} further reveals that both methods
failed completely across all tasks under out-of-distribution conditions, highlighting their limitations in generalization of precision manipulation.
SPOT employs relative pose as observation input, achieving a comparable success rate of 71.7\% under standard conditions, and maintaining 61.7\% average success rate when handling out-of-distribution cases. SPOT achieves  the same performance as the PoseDP w/ MLP, shown in Table \ref{ablation_in} and  Table \ref{ablation_out}. In contrast,  the PoseDP with disentangled pose encoder demonstrates superior performance, achieving an 81.7\% average success rate under standard conditions and maintaining 76.7\% average success rate with out-of-distribution cases, outperforming SPOT in both scenarios. 

We observed that ACT learned the downward insertion trend but failed to refine the hole-searching trajectory, even with wrist-camera feedback in task $2$. This suggests its visual encoder struggles to extract features relevant to the fine-grained trajectory. The absence of point cloud data prevents RISE from accurately localizing the contact region. Consequently, the model fails to model the fine-grained trajectory. Compared to ACT, RISE exhibits a broader action distribution and succeeds twice in task 1. 
  

Moreover, maintaining consistent source object pose relative to the gripper is challenging, both during data collection and inference. This inherent variability in contact-rich manipulation tasks frequently leads to pose shifts, posing significant challenges to the algorithm's generalization. In summary, both ACT and RISE fail to accurately model the search trajectory.

\subsection{Gated RGBD-Augmented Pose Diffusion Policy}

As shown in Table \ref{in_d} and Table \ref{out_d}, while PoseDP achieves high accuracy in tasks $1$ to $5$, its performance significantly degrades in task $6$ due to the small size of the USB (Type-C) connector and micro insertion clearance. This result demonstrates that the excessive pose estimation noise prevents PoseDP from learning complex insertion trajectories.
In contrast, the proposed gated RGBD-augmented pose diffusion policy (RPDP) attains a 100\% success rate under standard conditions and maintains an 80\% success rate even in out-of-distribution scenarios.

As shown in Fig. \ref{vis},  the training and test trajectories of task 6 are visualized. The source object undergoes insertion along the target object's z-axis. It is evident that the training trajectories exhibit significant noise, making it challenging for the PoseDP to accurately model the insertion trajectory. However, the proposed RPDP incorporates RGBD images to aid pose features, enabling the algorithm to model complex insertion trajectories from noisy training data.  In addition, we also show the position of the source object in the target object frame after insertion in the training/test set. The comparative analysis of train/test trajectories and post-insertion relative position (Fig.~\ref{vis}) reveals a pronounced divergence in the noise distributions of estimated poses between the train/test datasets. This demonstrates RPDP's high robustness to variations in pose estimation noise. In out-of-distribution cases, we observed that when the USB port becomes occluded, the policy fails more often, leading to degraded success rates. Under standard conditions, for the tasks $1$ to $5$, RPDP achieves a success rate similar to that of the PoseDP. This occurs because these tasks are less affected by noise (Fig.~\ref{data}(a)), making the auxiliary visual features less impactful. Furthermore, performance degrades in out-of-distribution cases due to limited generalization capability of visual features.

\subsection{Ablation Study}
In this section, we perform ablation studies for the DPE, GIE and PRGF.

As shown in Table \ref{ablation_in} and Table \ref{ablation_out}, PoseDP w/ MLP utilizes MLP to extract pose features; PoseDP w/  DPE utilizes disentangled pose encoder to extract pose features;  RPDP w/ Cat directly concatenates pose features and image features; RPDP w/ PRGF uses pose-guided residual gated fusion module to handle the pose features and image features.

\textbf{Disentangled Pose Encoder.} In Table \ref{ablation_in} and Table \ref{ablation_out}, the disentangled pose encoder achieves a higher success rate than  MLP encoder. The reason is that the DPE encodes rotation and translation separately for more accurate action prediction. Our experimental observations indicate that the use of the MLP encoder may induce jamming during the insertion process, consequently preventing full insertion.

\textbf{Goal-Conditioned RGBD Image Encoder.} In this study, the RGBD information is introduced to aid the  prediction of the complex search trajectory. As shown in Table \ref{ablation_in}, compared with PoseDP, while the direct concatenation of pose features and image features leads to performance degradation in tasks $1-5$, it achieves significant improvement in task $6$. This demonstrates that incorporating visual features enables the policy to model complex search trajectory despite noisy input conditions. For tasks $1-5$, the performance decline occurs because the high-dimensional image features dominate the low-dimensional pose representations.

With Pose-Guided Residual Gated Fusion (PRGF) module,  RPDP w/ PRGF shows a slight performance degradation in tasks $1-5$ compared to pose-only methods. This suggests that for these tasks, where pose observations alone provide sufficient information for successful completion, the introduction of visual features may cause potential overfitting. Conversely, RPDP w/ PRGF in task $6$ demonstrates significantly improved robustness. The USB (Type-C) insertion task is sensitive to pose estimation noise, making pose-only methods unable to learn meaningful insertion trajectories from noisy observations. In contrast, visual features provide complementary geometric constraints that enhance robustness.
  
\begin{figure}
	\centering
	\includegraphics[scale=0.45]{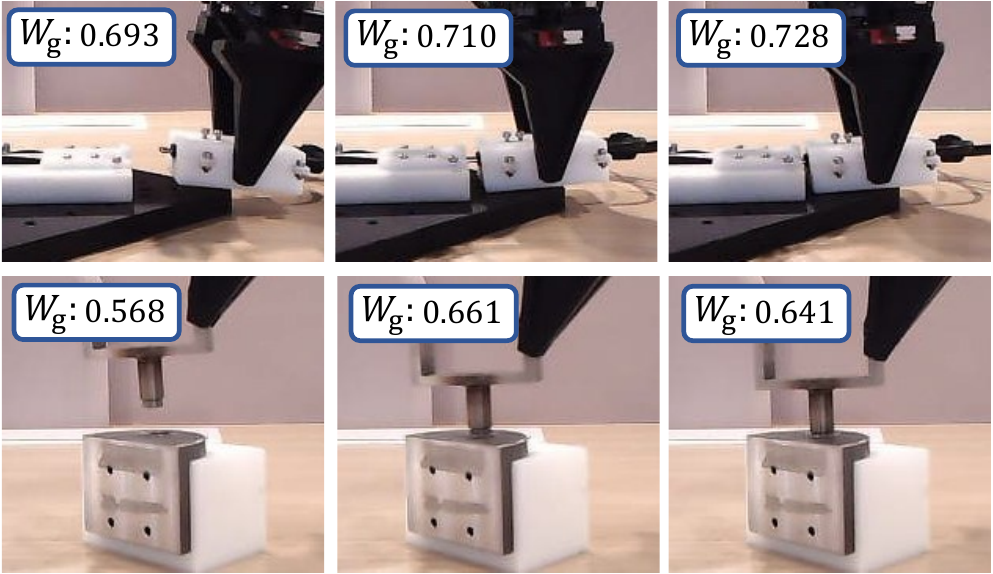}
		\vspace{-0.8em}
	\caption{The gating weights $W_g$ during the insertion process (Top: task $6$; Bottom: task $4$).
	}
	\label{gate}
\end{figure}

\textbf{Pose-Guided Residual Gated Fusion.} As shown in Table \ref{ablation_in} and Table \ref{ablation_out}, compared with directly concatenating pose features and image features,  RPDP w/ PRGF achieves a higher average success rate. The mean value of gating weights is computed, and shown in Fig.~\ref{gate}. The analysis of the gating weights reveals significantly higher visual feature utilization in task $6$ compared to task $4$, with progressive increase during contact/search phases. This demonstrates: (1) PRGF is capable of dynamically selecting visual features based on pose observation; (2) Visual features can enhance the ability of the policy to model complex insertion trajectories from noisy data.

\begin{table}[t]
	\caption{THE ABLATION STUDY OF OUR METHODS.}
	\centering
	\vspace{-0.8em}
	\label{ablation_in}
	\setlength{\tabcolsep}{1.8mm}{
		\begin{tabular}{cccccccc}
			\hline
			\textbf{Methods}   & \textbf{1}     & \textbf{2}    & \textbf{3}     & \textbf{ 4}     & \textbf{5}     & \textbf{6}     & \textbf{Average}   \\ \hline
			PoseDP w/ MLP      &\textbf{100} &70 &\textbf{90}  &90 &60 &20  &71.7  \\
			PoseDP w/ DPE      &\textbf{100} &\textbf{100} &\textbf{90}  &\textbf{100}  &\textbf{70} &30  &81.7  \\ \hline
			RPDP w/ Cat     &\textbf{100} &80 &60  &80  &30 &60  &68.3  \\
			RPDP w/ PRGF   &\textbf{100} &\textbf{100} &\textbf{90}  &90  &\textbf{70} &\textbf{100}  &\textbf{91.7}  \\
			\hline
	\end{tabular} }
	\vspace{-0.5em}
\end{table}

\begin{table}[t]
	\caption{THE ABLATION STUDY OF OUR METHODS. (OUT OF DISTRIBUTION)}
	\centering
	\vspace{-0.8em}
	\label{ablation_out}
	\setlength{\tabcolsep}{1.8mm}{
		\begin{tabular}{cccccccc}
			\hline
			\textbf{Methods}   & \textbf{1}     & \textbf{2}    & \textbf{3}     & \textbf{ 4}     & \textbf{5}     & \textbf{6}     & \textbf{Average}   \\ \hline
			PoseDP w/ MLP      &\textbf{90} &70 &80  &70  &40 &20  &61.7  \\
			PoseDP w/ DPE      &\textbf{90} &\textbf{100} &\textbf{90}  &\textbf{90}  &\textbf{60} &30  &76.7  \\ \hline
			RPDP w/ Cat     &70 &70 &40  &70  &20 &50  &53.3  \\
			RPDP w/ PRGF   &80 &90 &70 &\textbf{90}  &\textbf{60} &\textbf{80}  &\textbf{78.3}  \\
			\hline
	\end{tabular} }
	\vspace{-1.8em}
\end{table}

\subsection{Further Analysis}
\label{Further}
Here, we try to address the following research questions:

 \textbf{(1) Why can our methods accomplish precise insertion tasks despite hand–eye calibration error and robotic repeatability error?} Cobot Mobile ALOHA utilizes a low-cost robotic arm, whose repeatability is $1$ mm \cite{liu2024rdt}, and the hand-eye calibration error is about $2$ mm, which is included in $T_b^c$ (Alg. \ref{alg:pose_diffusion_policy}).
The repeatability error is included in $T_b^e$ (Alg. \ref{alg:pose_diffusion_policy}). However, interestingly, our methods are still capable of completing the insertion task with a clearance of $0.01$ mm. We argue that the proposed methods can model the complex trajectory during the insertion process, which is crucial for precise manipulation. In addition, the PoseDP can be considered as a pose-based servo controller, whereas RPDP integrates visual feedback with a pose-based servo controller, forming a hybrid framework. Furthermore, both PoseDP and RPDP only focus on the source and target objects, and both achieve closed-loop control.
 
\begin{figure}
	\centering
	\includegraphics[scale=0.27]{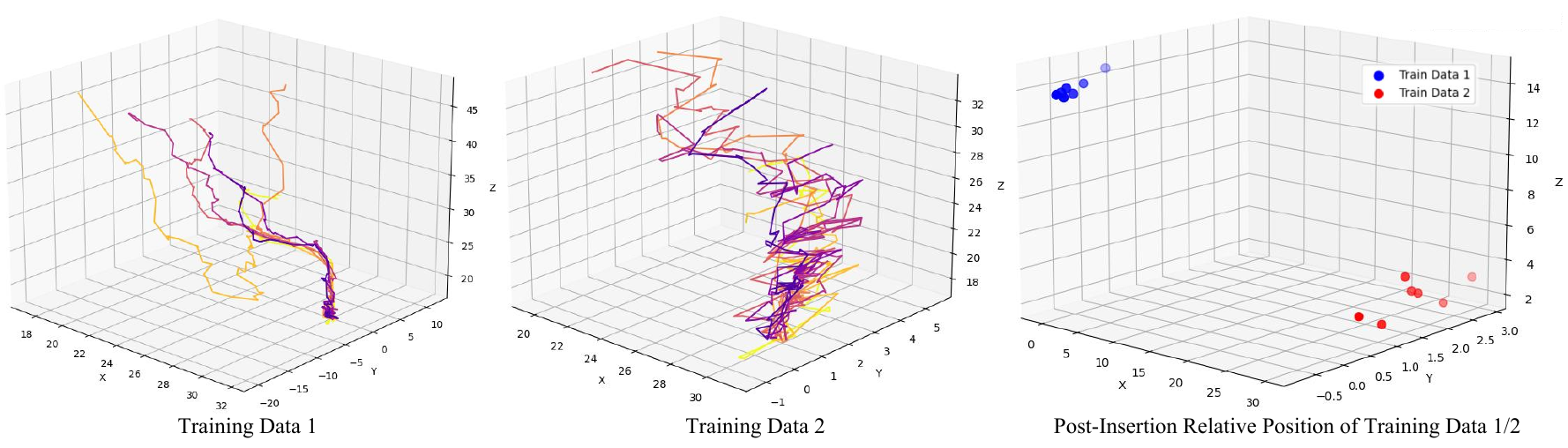}
			\vspace{-0.8em}
	\caption{Visualization of the trajectory and the post-insertion relative position of training data 1/2 (task $2$; Unit: Millimeter).
	}
	\vspace{-1.6em}
	\label{noise}
\end{figure}

\textbf{(2) How does the pose estimation noise affect the algorithm's performance?} 
Training data collected in the real-world inevitably contains noise. As illustrated in Fig.~\ref{noise}, we visualize the trajectories from training data 1/2, which exhibit significant divergence in post-insertion relative position.
It is obvious that training data 2 contains more noise. Our experiments reveal that the PoseDP trained on data 1 achieves a 100\% success rate, while that trained on data 2 only reaches a 60\% success rate. This demonstrates the importance of data quality for the algorithm's performance.

\textbf{(3) Runtime Analysis.} On an NVIDIA RTX 4060 GPU, PoseDP requires approximately 70 ms per prediction, while RPDP takes about 140 ms per prediction.

\subsection{Discussion and Limitation}
In this paper, we propose a pose-only diffusion policy (PoseDP) and a gated RGBD-augmented pose diffusion policy (RPDP) for robotic precise insertion. Unlike many prior RL-based insertion approaches that explicitly incorporate force/torque or tactile feedback, we focus on learning object-centric insertion trajectories using relative object pose (and pose-conditioned visual cues) as compact observations, and execute the policy on an ALOHA-like platform with passive compliance. Force/torque and tactile feedback are highly valuable for contact-rich manipulation, especially for rapid reactive behaviors and jamming recovery. However, policies that rely heavily on contact signals may become sensitive to the specific contact geometry and interaction conditions encountered during training, which can complicate generalization across pose variations. Our results suggest that predicting and executing object-centric trajectories in the relative pose space can provide a data-efficient alternative, while force/tactile sensing remains a complementary direction to further improve robustness in more challenging contact regimes.

Despite the promising results, our method has several limitations. First, it assumes that the source and target objects can be reliably estimated and tracked by a 6D pose estimator; for small or visually ambiguous objects (e.g., screws) pose estimation may be unreliable, and incorporating tactile/force feedback would likely be necessary. Second, pose tracking may fail under severe occlusions during manipulation. Employing additional viewpoints (e.g., a second camera) or active perception to maintain object visibility could mitigate this issue. Finally, our approach currently targets rigid objects and fixed object models; extending to broader object categories and improving robustness under large appearance changes remain important future directions.

\section{CONCLUSION}
In this paper, we explore pose-guided imitation learning for precise insertion tasks. The disentangled pose encoder is proposed to extract pose features. The goal-conditioned RGBD image encoder and the pose-guided residual gated fusion are proposed to augment pose features for complex trajectory modeling. We evaluate the proposed methods on six tasks and show that they significantly outperform current  imitation learning methods with impressive generalization. In addition, our methods only require $7-10$ demonstrations for each task. We hope this work encourages further study on integrating pose-centric policies with contact sensing for more challenging regimes.


\bibliographystyle{ieeetr}
\bibliography{my}   

@article{mnih2015human,
	title={Human-level control through deep reinforcement learning},
	author={Mnih, Volodymyr and  others},
	journal={nature},
	volume={518},
	number={7540},
	pages={529--533},
	year={2015},
	publisher={Nature Publishing Group UK London}
}

@article{chi2023diffusion,
	title={Diffusion policy: Visuomotor policy learning via action diffusion},
	author={Chi, Cheng and others},
	journal={The International Journal of Robotics Research},
	year={2023},
}

@INPROCEEDINGS{10801678,
	author={Wang, Chenxi and Fang, Hongjie and Fang, Hao-Shu and Lu, Cewu},
	booktitle={2024 IEEE/RSJ International Conference on Intelligent Robots and Systems (IROS)}, 
	title={RISE: 3D Perception Makes Real-World Robot Imitation Simple and Effective}, 
	year={2024},
	volume={},
	number={},
	pages={2870-2877}, }

@inproceedings{kim2024openvla,
	title={Open{VLA}: An Open-Source Vision-Language-Action Model},
	author={Moo Jin Kim and others},
	booktitle={Conference on Robot Learning (CoRL)},
	year={2024},
}

@inproceedings{mandlekar2023mimicgen,
	title={MimicGen: A Data Generation System for Scalable Robot Learning using Human Demonstrations},
	author={Mandlekar, Ajay and others},
	booktitle={Conference on Robot Learning  (CoRL)},
	year={2023}
}

@inproceedings{goyal2023rvt,
	title={Rvt: Robotic view transformer for 3d object manipulation},
	author={Goyal, Ankit and others},
	booktitle={Conference on Robot Learning (CoRL)},
	pages={694--710},
	year={2023},
}

@inproceedings{zhu2023viola,
	title={Viola: Imitation learning for vision-based manipulation with object proposal priors},
	author={Zhu, Yifeng and Joshi, Abhishek and Stone, Peter and Zhu, Yuke},
	booktitle={Conference on Robot Learning (CoRL)},
	pages={1199--1210},
	year={2023},
}

@INPROCEEDINGS{wen2022you, 
	AUTHOR    = {Bowen Wen and others}, 
	TITLE     = {{You Only Demonstrate Once: Category-Level Manipulation from Single Visual Demonstration}}, 
	BOOKTITLE = {Proceedings of Robotics: Science and Systems (RSS)}, 
	YEAR      = {2022}, 
}

@inproceedings{xu2024flow,
	title={Flow as the Cross-domain Manipulation Interface},
	author={Mengda Xu and Zhenjia Xu and Yinghao Xu and Cheng Chi and Gordon Wetzstein and Manuela Veloso and Shuran Song},
	booktitle={Conference on Robot Learning (CoRL)},
	year={2024},
}

@inproceedings{
	yuan2024general,
	title={General Flow as Foundation Affordance for Scalable Robot Learning},
	author={Chengbo Yuan and Chuan Wen and Tong Zhang and Yang Gao},
	booktitle={Conference on Robot Learning (CoRL)},
	year={2024},
}

@inproceedings{
	zhu2024vision,
	title={Vision-based Manipulation from Single Human Video with Open-World Object Graphs},
	author={Yifeng Zhu and Arisrei Lim and Peter Stone and Yuke Zhu},
	booktitle={Conference on Robot Learning Workshop on X-Embodiment Robot Learning},
	year={2024},
}

@INPROCEEDINGS{hsu2024spot,
	author={Hsu, Cheng-Chun and others},
	booktitle={2025 IEEE International Conference on Robotics and Automation (ICRA)}, 
	title={SPOT: SE(3) Pose Trajectory Diffusion for Object-Centric Manipulation}, 
	year={2025},
	pages={4853-4860},
}

@inproceedings{
	papagiannis2024miles,
	title={{MILES}: Making Imitation Learning Easy with Self-Supervision},
	author={Georgios Papagiannis and Edward Johns},
	booktitle={CoRL Workshop on Learning Robot Fine and Dexterous Manipulation: Perception and Control},
	year={2024},
}

@inproceedings{ze20243d,
	title={3d diffusion policy: Generalizable visuomotor policy learning via simple 3d representations},
	author={Ze, Yanjie and Zhang, Gu and Zhang, Kangning and Hu, Chenyuan and Wang, Muhan and Xu, Huazhe},
	booktitle={ICRA 2024 Workshop on 3D Visual Representations for Robot Manipulation},
	year={2024}
}

@inproceedings{shridhar2023perceiver,
	title={Perceiver-actor: A multi-task transformer for robotic manipulation},
	author={Shridhar, Mohit and Manuelli, Lucas and Fox, Dieter},
	booktitle={Conference on Robot Learning (CoRL)},
	pages={785--799},
	year={2023},
}

@inproceedings{thomas2018learning,
	title={Learning robotic assembly from cad},
	author={Thomas, Garrett and Chien, Melissa and Tamar, Aviv and Ojea, Juan Aparicio and Abbeel, Pieter},
	booktitle={2018 IEEE International Conference on Robotics and Automation (ICRA)},
	pages={3524--3531},
	year={2018},
}

@INPROCEEDINGS{10802423,
	author={Ota, Kei and others},
	booktitle={2024 IEEE/RSJ International Conference on Intelligent Robots and Systems (IROS)}, 
	title={Autonomous Robotic Assembly: From Part Singulation to Precise Assembly}, 
	year={2024},
	pages={13525-13532},
}

@article{mason2018toward,
	title={Toward robotic manipulation},
	author={Mason, Matthew T},
	journal={Annual Review of Control, Robotics, and Autonomous Systems},
	volume={1},
	number={1},
	pages={1--28},
	year={2018},
	publisher={Annual Reviews}
}

@inproceedings{wen2024foundationpose,
	title={Foundationpose: Unified 6d pose estimation and tracking of novel objects},
	author={Wen, Bowen and others},
	booktitle={Proc. IEEE Conf. Comput. Vis. Pattern Recognit. (CVPR)},
	pages={17868--17879},
	year={2024}
}

@inproceedings{dong2021tactile,
	title={Tactile-rl for insertion: Generalization to objects of unknown geometry},
	author={Dong, Siyuan and others},
	booktitle={2021 IEEE International Conference on Robotics and Automation (ICRA)},
	pages={6437--6443},
	year={2021},
}

@inproceedings{schoettler2020meta,
	title={Meta-reinforcement learning for robotic industrial insertion tasks},
	author={Schoettler, Gerrit and others},
	booktitle={2020 IEEE/RSJ International Conference on Intelligent Robots and Systems (IROS)},
	pages={9728--9735},
	year={2020},
}

@inproceedings{schoettlerdeep,
	title={Deep reinforcement learning for industrial insertion tasks with visual inputs and natural rewards},
	author={Schoettler, Gerrit and others},
	booktitle={2020 IEEE/RSJ International Conference on Intelligent Robots and Systems (IROS)},
	year={2020},
	pages={5548--5555}
}

@article{beltran2020learning,
	title={Learning force control for contact-rich manipulation tasks with rigid position-controlled robots},
	author={Beltran-Hernandez, Cristian Camilo and others},
	journal={IEEE Robotics and Automation Letters},
	volume={5},
	number={4},
	pages={5709--5716},
	year={2020},
}

@article{ma2024automated,
		title={Automated robotic assembly of shaft sleeve based on reinforcement learning},
		author={Ma, Xumiao and Xu, De},
		journal={The International Journal of Advanced Manufacturing Technology},
		volume={132},
		number={3},
		pages={1453--1463},
		year={2024},
		publisher={Springer}
}

@inproceedings{
	huang2024rekep,
	title={ReKep: Spatio-Temporal Reasoning of Relational Keypoint Constraints for Robotic Manipulation},
	author={Wenlong Huang and others},
	booktitle={Conference on Robot Learning (CoRL)},
	year={2024},
}

@inproceedings{
	song2020denoising,
	title={Denoising Diffusion Implicit Models},
	author={Jiaming Song and Chenlin Meng and Stefano Ermon},
	booktitle={International Conference on Learning Representations},
	year={2021},
}

@article{ho2020denoising,
title={Denoising diffusion probabilistic models},
author={Ho, Jonathan and Jain, Ajay and Abbeel, Pieter},
journal={Advances in neural information processing systems},
volume={33},
pages={6840--6851},
year={2020}
}

@INPROCEEDINGS{zhao2023learning,
	TITLE     ={Learning fine-grained bimanual manipulation with low-cost hardware},
	AUTHOR    = {Zhao, Tony Z and Kumar, Vikash and Levine, Sergey and Finn, Chelsea}, 
	BOOKTITLE = {Proceedings of Robotics: Science and Systems (RSS)}, 
	YEAR      = {2022}, 
}

@inproceedings{li2019cdpn,
	title={Cdpn: Coordinates-based disentangled pose network for real-time rgb-based 6-dof object pose estimation},
	author={Li, Zhigang and Wang, Gu and Ji, Xiangyang},
	booktitle={Proceedings of the IEEE/CVF international conference on computer vision},
	pages={7678--7687},
	year={2019}
}

@inproceedings{wang2021gdr,
	title={Gdr-net: Geometry-guided direct regression network for monocular 6d object pose estimation},
	author={Wang, Gu and others},
	booktitle={Proc. IEEE Conf. Comput. Vis. Pattern Recognit. (CVPR)},
	pages={16611--16621},
	year={2021}
}

@article{kim2024goal,
	title={Goal-conditioned dual-action imitation learning for dexterous dual-arm robot manipulation},
	author={Kim, Heecheol and Ohmura, Yoshiyuki and Kuniyoshi, Yasuo},
	journal={IEEE Transactions on Robotics},
	year={2024},
	publisher={IEEE}
}

@inproceedings{he2016deep,
	title={Deep residual learning for image recognition},
	author={He, Kaiming and Zhang, Xiangyu and Ren, Shaoqing and Sun, Jian},
	booktitle={Proceedings of the IEEE conference on computer vision and pattern recognition},
	pages={770--778},
	year={2016}
}

@inproceedings{dosovitskiy2020image,
	title	= {An Image is Worth 16x16 Words: Transformers for Image Recognition at Scale},
	author	= {Alexander Kolesnikov and others},
	booktitle={International Conference on Learning Representations},
	year={2021}, 		
}

@inproceedings{liu2024rdt,
	title={Rdt-1b: a diffusion foundation model for bimanual manipulation},
	author={Liu, Songming and others},
	booktitle={International Conference on Learning Representations (ICLR)},
	year={2025}, 	
}

\end{document}